%% file: Template_ISBI_latex.tex
\documentclass{article}
\usepackage{spconf,amsmath,graphicx}

\usepackage{enumitem}
\setlist{nosep, leftmargin=14pt}

\usepackage{mwe} 

\usepackage{graphicx}
\usepackage{amsmath}
\usepackage{amssymb}
\usepackage{booktabs}
\usepackage{bbm}

\usepackage{subfigure}
\usepackage{url}
\usepackage{diagbox}

\usepackage{xcolor}
\usepackage{breqn}
\usepackage{multicol}
\usepackage{multirow}

\DeclareMathOperator*{\diceavg}{D_{avg}}

\DeclareMathOperator*{\ged}{GED}
\DeclareMathOperator*{\dist}{D}
\DeclareMathOperator*{\samenc}{SAM_{enc}}
\DeclareMathOperator*{\samdec}{SAM_{dec}}

\DeclareMathOperator*{\iou}{IoU}

\DeclareMathOperator*{\convi}{conv-1}
\DeclareMathOperator*{\convii}{conv-2}

\title{SeqSAM: Autoregressive Multiple Hypothesis Prediction \\ for Medical Image Segmentation using SAM}
\name{Benjamin Towle$^{\star}$ \qquad Xin Chen$^{\star}$ \qquad Ke Zhou$^{\star \dagger}$}
\address{$^{\star}$ School of Computer Science, University of Nottingham \\  $^{\dagger}$ Nokia Bell Labs \\ \texttt{\{firstname.lastname\}@nottingham.ac.uk}}
\begin{document}
%
\maketitle
\begin{abstract}
\input{src/0_abstract}
\end{abstract}
\section{Introduction}
\label{sec:intro}
\input{src/1_introduction}

\section{Methodology}
\input{src/4_methodology}

\section{Experiments and Results}
\input{src/5_results}

\section{Conclusion and Future Work}
\input{src/6_conclusion}

\section{Acknowledgments}
\label{sec:acknowledgments}
This work is partly supported by the EPSRC DTP Studentship program. The opinions expressed in this paper are the authors’, and are not necessarily shared/endorsed by their employers and/or sponsors.

\section{Compliance with ethical standards}
This research study was conducted retrospectively using human subject data made available in open access. Ethical approval was not required as confirmed by the license attached with the open access data.

\bibliographystyle{IEEEbib}
\bibliography{strings,refs}

\end{document}

%% file: src/0_abstract.tex
Pre-trained segmentation models are a powerful and flexible tool for segmenting images. Recently, this trend has extended to medical imaging. Yet, often these methods only produce a single prediction for a given image, neglecting inherent uncertainty in medical images, due to unclear object boundaries and errors caused by the annotation tool. Multiple Choice Learning is a technique for generating multiple masks, through multiple learned prediction heads. However, this cannot readily be extended to producing more outputs than its initial pre-training hyperparameters, as the sparse, winner-takes-all loss function makes it easy for one prediction head to become overly dominant, thus not guaranteeing the clinical relevancy of each mask produced. We introduce \textsc{SeqSAM}, a sequential, RNN-inspired approach to generating multiple masks, which uses a bipartite matching loss for ensuring the clinical relevancy of each mask, and can produce an arbitrary number of masks. We show notable improvements in quality of each mask produced across two publicly available datasets. Our code is available at \url{https://github.com/BenjaminTowle/SeqSAM}.   


%% file: src/1_introduction.tex
Pre-trained image segmentation models are a powerful and flexible tool for segmenting images, without relying on domain-specific bespoke models \cite{sam}. Increasingly, these techniques are being applied to medical imaging, where the goal is to annotate the boundaries of an anatomical structure such as a tumour or lesion \cite{medsam}. However, medical images contain large amounts of uncertainty. This often leads to disagreement over the `correct' annotation for a given image, even amongst experts, due to confounding factors such as unclear boundaries of the structure of interest or errors introduced by the annotation tool used \cite{Baumgartner2019PHiSegCU, Monteiro2020StochasticSN}.

\begin{figure*}
    \centering
    \includegraphics[scale=0.1]{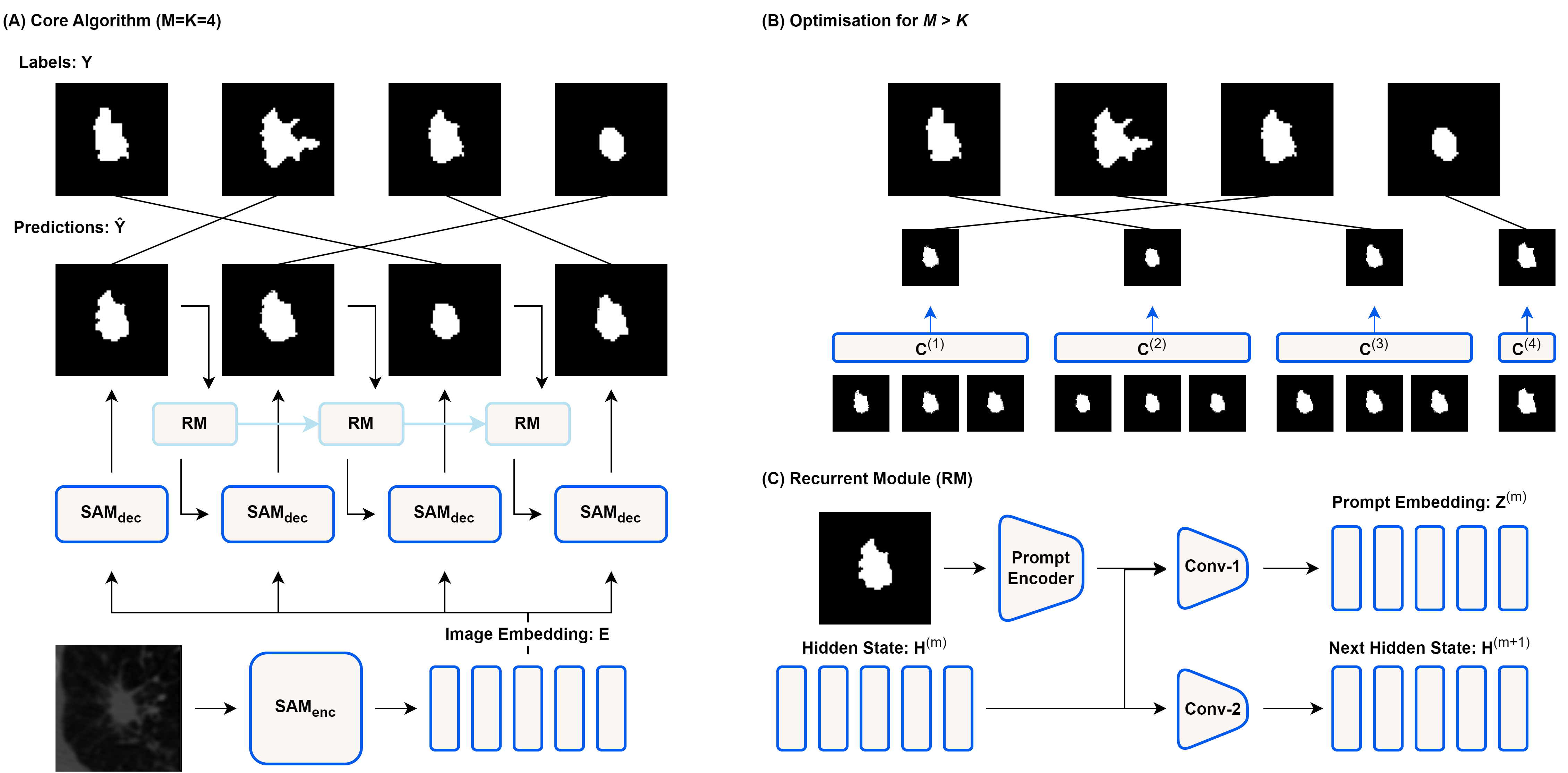}
    \caption{(A) Overview of \textsc{SeqSAM} comprising an encoder ($\samenc$) and a decoder ($\samdec$). (B) A Recurrent Module sequentially generates masks. (C) When more predictions than labels are generated, we chunk the predictions into groups where $C^{(k)}$ represents the $k-$th group and perform subsampling. Image and annotations are cropped to highlight region-of-interest.}
    \label{fig:seqsam}
\end{figure*}

To combat this, Multiple Choice Learning (MCL) was introduced as a method to generate multiple masks for a given image \cite{GuzmnRivera2012MultipleCL}, being most recently popularised as part of the SAM architecture, a large pre-trained segmentation model \cite{sam}. Specifically, under this approach, multiple prediction heads separately predict an image mask, and only the best image mask is backpropagated through. In medical imaging, the practical application of these techniques are numerous: MCL enables an intuitive and rapid form of semi-automatic annotation, by allowing a clinician to select from the most accurate annotation among a set of hypotheses; further, it is aligned with the collective intelligence-based approach reflected in the modern practice of case conferences and tumour boards \cite{Rahman2023AmbiguousMI}, which may alert a clinician to alternative hypotheses for a given segmentation. Indeed, the decision of where to draw the contour for a tumour can have significant medical consequences, such as in determining how invasive a surgery is \cite{Monteiro2020StochasticSN}. Yet, MCL has several challenges in a medical setting. In particular, MCL is limited to the fixed number of outputs it is trained on, which may fail to represent the full range of possible annotations, which often comprise subtle differences in contours. Further, the sparse winner-takes-all loss function makes it easy for one prediction head to become overly dominant, thus not guaranteeing the clinical relevancy of each mask produced.

Drawing inspiration from recurrent neural networks, we introduce \textsc{SeqSAM}, which sequentially generates multiple masks via a single prediction head. We further implement a set-based optimisation procedure to enable credit assignment and ensure the clinical relevancy of each mask. Finally, we extend our approach to an arbitrary number of masks using randomised strided pooling. We evaluate on two popular multi-annotator datasets: LIDC-IDRI \cite{lidc} and QUBIQ Kidney \cite{qubiq}, and demonstrate notable improvements in quality the multiple masks produced. In summary, our main contributions are: (1) We introduce an autoregressive framework to the multiple mask generation task. (2) We show the novel method \textsc{SeqSAM}, which transforms SAM into an RNN-like architecture for generating multiple masks, using a novel randomised strided pooling approach for generating an arbitrary number of masks. (3) We present comprehensive empirical evaluation of our approach across accuracy and distribution metrics, and provide a detailed ablation study of our method.

%% file: src/4_methodology.tex

\subsection{Task Definition.} We begin with the dataset $\mathcal{D} = \{\langle \textbf{x}_i,\mathbf{Y}_i \rangle \}^I$ where the $i$-th sample consists of an image of $J$ pixels $x_i^{(j)} \in \textbf{x}_i$ and a set of $K$ plausible binary masks $\textbf{y}_i^{(k)} \in \mathbf{Y}_i$. Our goal is to learn the function that maps the input to $M$ outputs, where the prediction for the $i$-th sample is defined as the set of $M$ hypotheses $\mathbf{\hat{y}}_i = \{\hat{\textbf{y}}_i^{(1)},...,\hat{\textbf{y}}_i^{(M)}\}$. Note that $M$ does not have to equal $K$, and we demonstrate approaches for dealing with $M != K$ in Section \ref{sec:m!=k}.

\subsection{Segment Anything Model.} SAM \cite{sam} -- the backbone of our method and baseline comparison -- is a transformer-based segmentation model pre-trained on over 1 billion image masks from 11M images. The model enables various forms of user interaction such as bounding boxes, clicks and masks (sc. `prompts') and comprises a large image encoder $\samenc$, with a lightweight mask decoder $\samdec$ and prompt encoder. Because the interaction information is only injected at a late stage in the model pipeline, multiple masks can be generated and adapted rapidly by changing the interaction inputs to the decoder, without requiring the original image to be re-encoded. At the final output layer of the model is either a single prediction head or a multi-output prediction head, depending on whether the model is being used to generate a single or multiple outputs. For the MCL baseline, we use the multi-output prediction head, while we use the single-prediction head for our method, as multiple masks can be generated through prompting the model with different input masks. We note that for both our approach and the baseline approaches, the latency is roughly comparable, as most computation is taken up by the encoder which comprises the vast majority of the model's parameters.

\subsection{Recurrent Module.} By default, SAM encodes image $\textbf{x}$ to obtain the embedding $\mathbf{E} \in \mathbb{R}^{C \times H \times W}$, which the decoder then transforms into a single logits mask via its single-prediction head: $\textbf{z}^{(1)}$. By leveraging SAM's native prompt encoder we can generate the next mask, simply by feeding $\textbf{z}^{(1)}$ as an auxiliary input into the mask decoder. However, this is a stateless process, and would prevent SAM from tracking long-term information about which masks have been generated. Therefore, we connect this process via a Recurrent Module \cite{rnn}. Specifically, we define a hidden state $\mathbf{H}^{(m)}$, initialised as zeros for $m = 1$, which maintains a memory of previously seen masks. At each timestep, we update this hidden state by applying a single convolutional layer over the channel-wise concatenation of the last hidden state and current logits mask: $ \mathbf{H}^{(m+1)} = \convii ([ \mathbf{H}^{(m)} , \mathbf{Z}^{(m)}])$. In parallel, we apply a separate convolution over the same inputs to provide a sequence-aware prompt embedding for the SAM decoder:
$\mathbf{Z}^{(m)+} = \convi ([ \mathbf{H}^{(m)} , \mathbf{Z}^{(m)}])$, as shown in Figure \ref{fig:seqsam}.

\subsection{Set-based Optimisation.} By unrolling over $M$ timesteps, we obtain a sequence of logits masks $\{\mathbf{z}^{(1)},...,\mathbf{z}^{(M)}\}$ which can be mapped into probabilities through a pixel-wise sigmoid function. However, unlike in sequence prediction tasks such as language modeling \cite{rnn}, our label set $\mathbf{Y}$ lacks canonical ordering. Therefore, naively assigning the $k$-th label to the $m$-th predicted mask is unlikely to be effective for training, as the model does not know what order the labels will appear in. This is broadly referred to as the credit-assignment problem \cite{Zhang2019DeepSP}.  To mitigate this, we treat the mapping between predictions and labels as a linear assignment problem. In particular, let $\mathbf{Y}_\pi$ represent the $\pi$-th permutation of labels, from a space of $\Pi$ possible permutations. Then we define our loss function as:
\begin{equation}
\label{eq:loss}
    \mathcal{L}(\mathcal{\hat{\mathbf{Y}}}, \mathcal{\mathbf{Y}}) = \min_{\pi=1:\Pi} \sum_{m=1}^M \mathcal{L} (\hat{\mathbf{y}}^{(m)}, \mathbf{y}_{\pi}^{(m)})
\end{equation}
where $M$ equals the $K$ available labels. This can be efficiently computed in $O(N^3)$ using the Hungarian algorithm \cite{detr}. As each prediction is assigned to exactly one label, it has the advantage of enforcing that the model captures the range of possible modes represented in the labels. Due to the sequential nature of the model, it is encouraged to canonicalise its generated masks. For our choice of loss function, we use dice loss, following previous work \cite{medsam}. Note that these logits are backpropagated through time during training, encouraging the logits to track information about previous masks predicted in the sequence.


\subsection{Generating Different Numbers of Masks.} 
\label{sec:m!=k} The downstream inference use case may require the model to generate a different number of masks $M$ than the number of labels ($K$) per sample in the dataset. When the number of masks to be generated is less than the number of labels, a straightforward approach we employ is to simply train using the available number of labels, and then to truncate the generation during inference.

However, when the number of masks to be generated is larger than the available number of labels, forcing the model to generate additional masks beyond the sequence length it was trained on may lead to out-of-distribution issues. 

To mitigate this, we adopt a randomised strided pooling approach, as shown in Figure \ref{fig:seqsam}B. In particular, we first generate a sequence of logits $\{\mathbf{z}^{(1)},...,\mathbf{z}^{(M)}\}$ as before, equal to the desired sequence length $M$. We then divide the sequence into $K$ chunks where each chunk is comprised of neighbouring masks from the sequence. Specifically, each chunk $\mathbf{C}^{(k)} \in \{\mathbf{C}^{(1)},...,\mathbf{C}^{(K-1)} \}$ has the size $\lceil \frac{M}{K} \rceil$, and the final chunk $\mathbf{C}^{(K)}$ has the size $M \mod K$. During training, we then randomly sample a mask from each chunk, to obtain $K$ masks, and perform the same optimisation as before (Equation (\ref{eq:loss})). During inference, we omit the sampling process, and use all of generated masks. Intuitively, this encourages better credit assignment for the model, as each chunk can become specialised for a particular role.

%% file: src/5_results.tex
\begin{table}[t]
\centering
\setlength{\tabcolsep}{2.5pt}
\footnotesize
    \include{src/simsr_results3}
    \caption{Results on LIDC-IDRI and QUBIC Kidney test sets. (A) Performance of SeqSAM versus SAM MCL baseline. \textbf{Bold} indicates best result. $\dag$ indicates the result is significantly better than the baseline on Wilcoxon Signed Rank Test ($p < 0.01$). (B) Effect of various ablations on \textsc{SeqSAM}.}
    \label{tab:main_results}
\end{table}

\begin{figure*}
    \centering
    \begin{subfigure}
    \centering
        \includegraphics[scale=0.35]{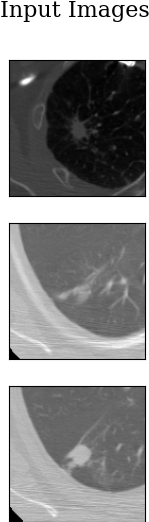}
    \end{subfigure}
    \begin{subfigure}
    \centering
        \includegraphics[scale=0.35]{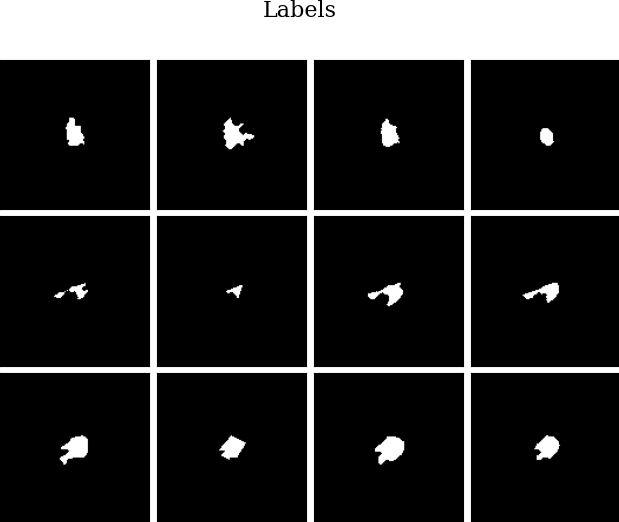}
    \end{subfigure}
    \begin{subfigure}
    \centering
        \includegraphics[scale=0.35]{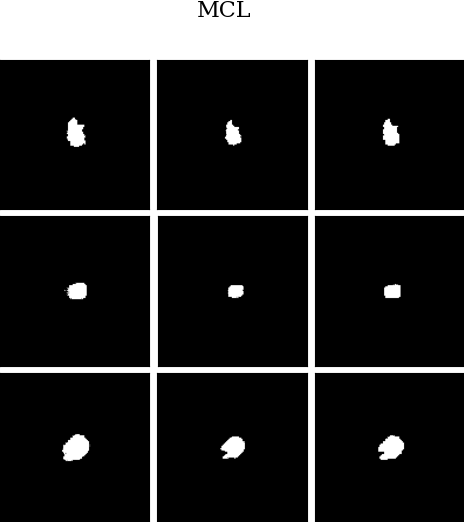}
    \end{subfigure}
    \begin{subfigure}
    \centering
        \includegraphics[scale=0.35]{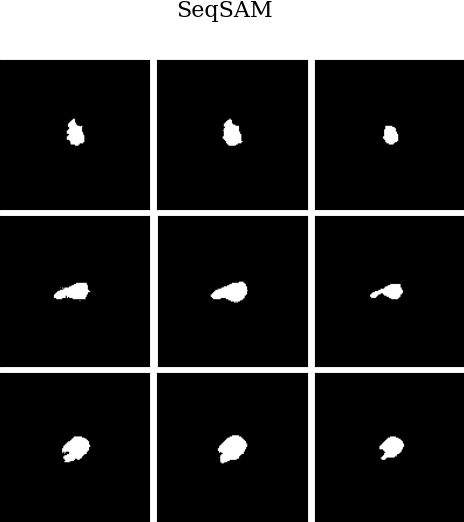}
    \end{subfigure}
    \caption{Qualitative results on LIDC-IDRI test set using SAM, showing annotated lung nodules. (Col. 1) Input images; (Col. 2-5) labels from multiple human annotators; (Col. 6-8) MCL baseline \cite{sam}; (Col. 9-11) \textsc{SeqSAM}, our method, for $M = 3$.}
     \label{fig:case_study}
\end{figure*}

\subsection{Experimental Design}
\noindent
\textbf{Datasets.} We evaluate on two multi-annotator datasets: (i) Lung Image Database Consortium image dataset (LIDC-IDRI) \cite{lidc} ($K = 4$) and (ii) QUBIQ Kidney \cite{qubiq}. LIDC-IDRI comprises 1018 3D thorax CT scans ($K = 3$). We use the preprocessed version provided by Kohl et al. \cite{Kohl2018APU}, which processes the data into 2D slices centred around each lung nodule. This results in a dataset of 15,096 2D images. We randomly split the dataset 80\% / 10\% / 10\% for training, validation and testing respectively. QUBIQ Kidney contains 24 CT scans with 3 separate annotations; the task is to annotate the contours of the kidney and we use the default train / test split of 20 / 4 respectively. Note that one limitation of our approach is that it requires a dataset with multiple annotations per image, which are much rarer than single annotator datasets. Hence, we only use the aforementioned two datasets.


\noindent
\textbf{Model Training and Parameter Setting.} We train with a batch size of 2 using the AdamW optimiser \cite{adamw}, a learning rate of 1e-4 and weight decay of 0.01, until convergence. We froze the image encoder, to limit the number of parameters required to be updated. As SAM is designed for semi-automatic segmentation, for each image, we provided a bounding box, which previous studies found to be the most effective form of prompting \cite{medsam}. As each image contains multiple annotations, we achieve this by using the extremity points of a randomly selected non-empty mask from each label set. We baseline our method against a state-of-the-art (SoTA) MCL approach \cite{sam}. We also compare using MedSAM, a deterministic variant of SAM specialised on medical images \cite{medsam}, which can also be extended to the \textsc{SeqSAM} method.

\noindent
\textbf{Metrics.} We evaluate on both accuracy and distribution metrics. First, dice score -- $\diceavg$ -- computes the average similarity across all predictions and labels, from first aggregating the predictions by pixel-wise majority voting, and then by averaging over dice scores with each label. Then, Generalised Energy Distance (GED) is used to measure whether the generated masks capture the full \textit{range} of possible modes for each image: 




\begin{dmath}
\label{eq:ged}
    \ged = 2\mathbb{E}_{\mathbf{y} \sim \mathbf{Y},\hat{\mathbf{y}} \sim \mathbf{\hat{Y}}} [\dist(\mathbf{y},\hat{\mathbf{y}})] - \mathbb{E}_{\mathbf{y} \sim \mathbf{Y},\mathbf{y}' \sim \mathbf{Y}'} [\dist(\mathbf{y},\mathbf{y}')] - \mathbb{E}_{\hat{\mathbf{y}} \sim \mathcal{\hat{\mathbf{Y}}}, \hat{\mathbf{y}}' \sim \mathcal{\hat{\mathbf{Y}}}'} [\dist(\hat{\mathbf{y}},\hat{\mathbf{y}}')] 
\end{dmath}
where $\dist (\cdot,\cdot)$ is a distance function given as $1 - \iou(\cdot,\cdot)$ and $\iou$ (intersection-over-union) is given as: $\iou (\hat{\mathbf{y}}, \mathbf{y}) = \frac{\lvert \hat{\mathbf{y}} \cap \mathbf{y} \rvert}{\lvert \hat{\mathbf{y}} \cup \mathbf{y} \rvert}$. Note that for deterministic approaches $\dist (\mathbf{y},\mathbf{y}')$ = 0.


\subsection{Results}
\noindent
\textbf{Main Results.} \label{sec:main_results} Table \ref{tab:main_results}A compares our method to the (Med)SAM baselines. When generating the same number of masks, our approach produces masks that are often more accurate, and also better capture the distribution of labels. We further validated the idea that our approach is able to generate an arbitrary numbers of masks. When setting the number of masks to 10, our model still produces masks that are equally as clinically relevant as when the number of masks is 3, demonstrating the flexibility of its use cases. This reveals that the additional masks are not simply copies of earlier masks but in fact capture new plausible annotations. Note statistical significance is harder to show for QUBIQ Kidney due to the smaller dataset size, although relative ranking of methods is aligned with LIDC-IDRI.

\noindent
\label{sec:ablation}
\textbf{Ablation Study.} Table \ref{tab:main_results}B motivates our design choices with several ablations. In particular, we explore different ways of selecting the $K$ masks during training, when the inference number of masks $M$ is larger. We first use a random masks approach, whereby instead of splitting the $M$ masks into chunks first, we sample purely from the sequence of $M$ masks, thus making the sequence non-canonical. Then, we explore just taking the first $K$ masks from the sequence. We find model accuracy deteriorates slightly when using random masks, suggesting the model struggles with credit assignment. When taking only the first $K$ masks, we find very large declines in accuracy, suggesting it is important for the sequence length the model is exposed to during training to align with the number of masks produced during inference. Finally, we explored removing the backpropagation-through-time from the model's logits -- in this case, backpropagation only occurs through the Recurrent Module's hidden state. We found this also reduced performance, indicating that the model's logits were also being used to effectively track information about previous masks through time.  

\label{sec:case_study}

\noindent
\textbf{Qualitative Analysis.} Figure \ref{fig:case_study} provides qualitative examples of our model from the LIDC-IDRI test set. The labels reveal significant annotator disagreement regarding size and contours of the object of interest. However, the baseline MCL often misses this disagreement, also producing annotations that are more coarse-grained and simplistic in nature. By contrast, our approach, \textsc{SeqSAM}, is able to capture more complexity and nuance in the shape, as well as highlighting points of disagreement also present in the human annotators.

%% file: src/simsr_results3.tex
\definecolor{blue}{HTML}{025BED}
\begin{tabular}{lllcllll}
\toprule
& &  & & \multicolumn{2}{c}{LIDC-IDRI} & \multicolumn{2}{c}{QUBIQ Kidney} \\
\cmidrule(lr){5-6} \cmidrule(lr){7-8}
\colorbox{blue}{\textcolor{white}{A}} & Base & Method & $M$ & $\diceavg ^\uparrow$ & $\ged ^\downarrow$ &  $\diceavg ^\uparrow$ & $\ged ^\downarrow$ \\
\midrule
& \multirow{3}{*}{SAM \cite{sam}} & MCL & 3 & 82.4 & .284 & 89.2 & .257 \\
& & \textsc{SeqSAM} & 3 & \textbf{83.8} $\dag$ & .245 $\dag$ & \textbf{90.0} & .241 \\
& & \textsc{SeqSAM} & 10 & 83.5 & \textbf{.234} $\dag$ & 89.4 & \textbf{.235} \\
\midrule
& \multirow{3}{*}{MedSAM \cite{medsam}} & Deterministic & 1 & 84.0 & .344 & 89.3 & .297 \\
& & SeqSAM & 3 & 83.9 & .248 $\dag$ & 89.3 & .265 \\
& & SeqSAM & 10 & \textbf{84.2} & \textbf{.213} $\dag$ & \textbf{89.9} & \textbf{.210} \\
\midrule
\midrule
\colorbox{blue}{\textcolor{white}{B}} & \multicolumn{7}{c}{Ablations} \\
\midrule
& \multirow{3}{*}{SAM \cite{sam}} & Random $K$ & 10 & 83.0 & .237 & 87.6 & .285  \\
& & First $K$ & 10 & 78.7 & .308 & 87.7 & .238 \\
& & No Backprop. & 3 & 83.2 & .255 & 89.8 & .236 \\  
\bottomrule
\end{tabular}

%% file: src/6_conclusion.tex
We introduce \textsc{SeqSAM}, a novel method for generating multiple clinically relevant masks, using a novel autoregressive mask generation algorithm. We show SoTA performance across two multi-annotator datasets, while combining superior flexibility to previous methods through being trained to generate an arbitrary number of predictions.

